\documentclass[runningheads]{llncs}

\usepackage[mobile]{eccv}

\definecolor{citecolor}{HTML}{2980b9}
\usepackage{eccvabbrv}
\usepackage{algorithm,algpseudocode}
\usepackage{graphicx}
\usepackage{booktabs}
\usepackage{multirow}
\usepackage{makecell}
\usepackage{caption}
\usepackage{float}
\usepackage{wrapfig}
\usepackage{multirow}
\usepackage[accsupp]{axessibility}  
\usepackage{adjustbox}
\usepackage{framed}
\usepackage{makecell}
\usepackage{listings}

\usepackage{tabularx} 
\usepackage{geometry}  

\usepackage{color, colortbl}
\usepackage{orcidlink}
\newcommand{\name}{DetToolChain }
\usepackage{tablefootnote}
\begin{document}

\title{DetToolChain: A New Prompting Paradigm to Unleash Detection Ability of MLLM} 

\titlerunning{DetToolChain: Chain-of-Thought for Detection}

\author{Yixuan Wu\inst{125}  \and
Yizhou Wang\inst{2}  \and
Shixiang Tang\inst{3}  \and
Wenhao Wu\inst{4}  \and
Tong He\inst{2}  \and\\
Wanli Ouyang\inst{2}  \and
Philip Torr\inst{5}   \and
Jian Wu\inst{1} 
}

\authorrunning{Y. Wu et al.}


\institute{$^{1}$ Zhejiang University
$^{2}$ Shanghai AI lab \\$^{3}$ The Chinese University of Hong Kong   \\$^{4}$ The University of Sydney $^{5}$  University of Oxford}

\maketitle

\begin{abstract}
We present DetToolChain, a novel prompting paradigm, to unleash the zero-shot object detection ability of multimodal large language models (MLLMs), such as GPT-4V and Gemini. Our approach consists of a detection prompting toolkit inspired by high-precision detection priors and a new Chain-of-Thought to implement these prompts. Specifically, the prompts in the toolkit are designed to guide the MLLM to focus on regional information (e.g., zooming in), read coordinates according to measure standards (e.g., overlaying rulers and compasses), and infer from the contextual information (e.g., overlaying scene graphs). Building upon these tools, the new detection chain-of-thought can automatically decompose the task into simple subtasks, diagnose the predictions, and plan for progressive box refinements.
The effectiveness of our framework is demonstrated across a spectrum of detection tasks, especially hard cases. Compared to existing state-of-the-art methods, GPT-4V with our DetToolChain improves state-of-the-art object detectors by +\textbf{21.5\%} $AP_{50}$ on MS COCO \texttt{Novel class} set for open-vocabulary detection, \textbf{+24.23\%} $Acc$ on RefCOCO val set for zero-shot referring expression comprehension, \textbf{+14.5\%} $AP$ on D-cube describe object detection FULL setting. 
\end{abstract}

\section{Introduction}
\label{sec:intro}
Large language models (LLMs), \emph{e.g.}, GPT3~\cite{brown2020language}, Gemini~\cite{team2023gemini}, InternLM~\cite{team2023internlm}, and Qwen~\cite{bai2023qwen}, have shown unprecedented capabilities in understanding human languages and solving practical problems such as scientific question answering and code generation. When integrated with visual encoders, large language models can be upgraded to multimodal large language models (MLLMs), which can achieve the ability similar to human visual intelligence and tackle some visual understanding tasks, such as image captioning. Despite these advances, the potential of MLLMs in detection tasks is still underestimated among common vision tasks~\cite{yang2023dawn, yin2024lamm,zang2023contextual,tong2024eyes}. When required to ask precise coordinates in complicated object detection tasks, \emph{e.g.,} detecting highly occluded objects, rotational objects, or small objects in the scene images, MLLMs often miss the target objects or answer inaccurate bounding boxs~\cite{yang2023dawn}. The poor performance on object detection significantly limits the applications of MLLMs in the real world, \emph{e.g.,} defect detection~\cite{8960347,zeng2022small,czimmermann2020visual} and sports analysis~\cite{thomas2017computer,buric2018object,vandeghen2022semi}.

To enhance the detection capabilities of MLLMs, prior efforts can be categorized into two classes: (1) Finetuning MLLMs with high-quality question-answer instructions with abundant location information~\cite{chen2023shikra,lin2023sphinx,Bai2023QwenVLAF,peng2023kosmos,lin2023sphinx} in the answer. Despite the considerable improvements achieved, preparing high-quality question-answer pairs requires great manual efforts and finetuning multimodal large language models suffers from large computational costs. Furthermore, since current state-of-the-art MLLMs~\cite{team2023gemini,Bai2023QwenVLAF,achiam2023gpt} are closed-sourced and their performances have been significantly superior to open-sourced models, the ``finetuning'' method can not be implemented on the most powerful MLLMs at the moment (and most likely in the future), which significantly limits its potential to continuously improve the emerging state-of-the-art MLLMs. (2) Designing textual or visual prompting with location information to advance the localization ability of MLLMs. While intuition-based prompting methods have greatly advanced the performance of regional comprehension tasks such as compositional reasoning~\cite{mitra2023compositional} and spatial understanding~\cite{chen2024spatialvlm}, their effectiveness on detection tasks remains underexplored.

This work explores how the detection ability of multimodal large language models can be unlocked by a new chain of thoughts on detection prompting toolkits (dubbed as DetToolchain). The new DetToolchain is motivated by three ideas. 
First, visual prompts are identified as a crucial component of the detection prompting toolkits. They offer a more direct and intuitive approach to enhancing the spatial comprehension of Multimodal Large Language Models (MLLMs) compared to language prompts. 
This is because current MLLMs still struggle with accurately translating textual coordinates and descriptions into precise regions and visual information.
Instead, visual prompts directly drawn in the image can significantly narrow the gap between visual and textual information and ultimately contribute to the improved detection ability of MLLMs.
Second, detecting challenging instances, such as occluded and small objects, can be more efficiently tackled by breaking them down into smaller, simpler subtasks.
Third, the detection results should be modified step by step using Chain-of-Thought, similar to the progressive refinement of bounding boxes in current state-of-the-art object detection algorithms such as DETR~\cite{carion2020end}, SparseRCNN~\cite{sun2021sparse}, and DiffusionDet~\cite{chen2023diffusiondet}. 
Based on these ideas, our proposed DetToolChain consists of a detection prompts toolkit, including \emph{visual processing} prompts and \emph{detection reasoning} prompts, and a multimodal Chain-of-Thought to properly apply these detection prompts to unleash the detection ability of MLLMs. Their critical designs and insights involve:

\begin{figure}[t]
  \centering
  \includegraphics[height=5.4cm]{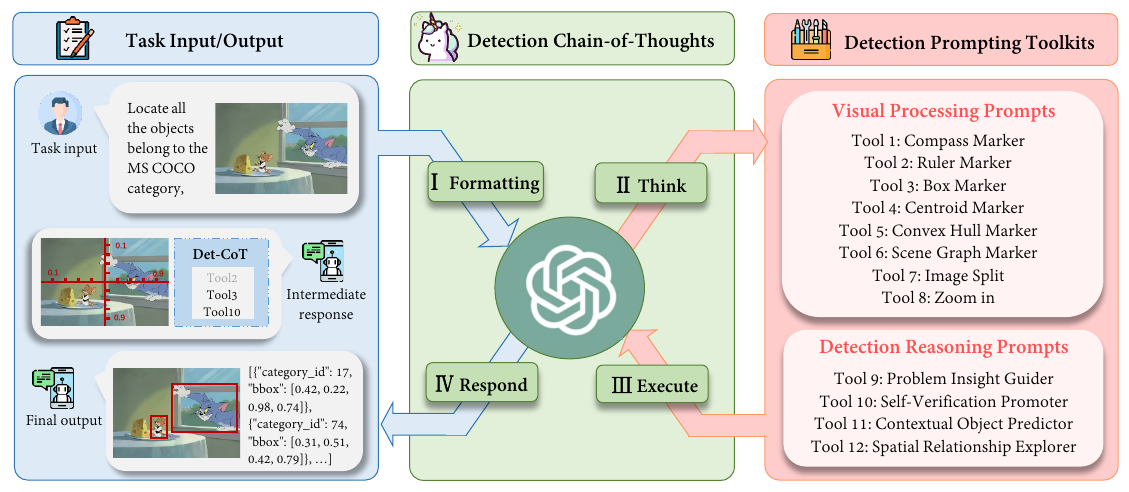}
  \caption{
The overall framework of DetToolChain for unleashing the detection potential of MLLMs. \emph{e.g.}, GPT-4V. Given a query image, the MLLM is instructed to: \uppercase\expandafter{\romannumeral1}. format the task raw input into a suitable instruction template, \uppercase\expandafter{\romannumeral2}. decompose the specific detection task into simpler subtasks and choose effective prompts from Detection Prompting Toolkits, \uppercase\expandafter{\romannumeral3}. execute the specific prompt sequentially, \uppercase\expandafter{\romannumeral4}. after verifying the predictions, return the response.
 }
  \label{fig:framework}
  \vspace{-2em}
\end{figure}

\textbf{(1) A comprehensive set of \emph{visual processing} prompts that support a wide range of detection tasks.} 
The \emph{visual processing} prompts process the given images to facilitate the detection performance of MLLMs, according to some well-accepted prior knowledge and techniques in effective detectors. Specifically, the \emph{visual processing} prompts can be divided into three categories, \emph{i.e.,} the regional amplifier, the spatial measurement standard, and the scene image parser. These prompts facilitate different key factors for better detectors. First, the regional amplifier consists of image splitting and zooming in, capable of highlighting the region of interest in detection tasks. Second, the spatial measurement standard includes rulers and compass with linear graduations, which provide transnational references and rotational references in object detection (particularly in rotated object detection), respectively. Similar to human intelligence, these spatial measurement standard can help MLLMs to locate the objects and read their coordinates out. Third, the scene parser marks the predicted positions or spatial relations of objects in the images by convex hull, bounding boxes of objects and scene graphs of the image. The markers by the scene parser can facilitate the detection capability of MLLMs by encouraging them to reason from contextual information in the scene images.

\textbf{(2) A comprehensive set of \emph{detection reasoning} prompts that help MLLMs to diagnose the detection results and reason the next \emph{visual processing} prompts to be applied.} Different from \emph{visual processing} prompts, the \emph{detection reasoning} prompts do not process the images but are dedicated to evaluating the predicted bounding boxes and diagnosing the inaccurate predictions even using the \emph{visual processing} prompts.
By analyzing the relationship and reasoning the co-occurrence of detected objects in the scene image with the commonsense knowledge in MLLM, these prompts help the MLLM to prevent hallucination on detection tasks.

\textbf{(3) A multimodal detection Chain-of-Thought (Det-CoT) that enables the MLLM to manage the whole process of detecting targets.} 
With these powerful detection prompting toolkits, the detection Chain-of-Thought helps the MLLM comprehend spatial detection tasks and produce reliable bounding boxes. As illustrated in Fig.~\ref{fig:framework}, the multimodal detection Chain-of-Thought are instructed to (1) format the raw input with a suitable instruction template, (2) decompose the complex detection task into smaller sub-tasks and select the corresponding detection prompting toolkit, (3) execute detection prompting toolkit iteratively, and (4) apply its own reasoning and critical thinking to oversee the whole detection process and return the final response. Given a detection task, our proposed Det-CoT can automatically decompose it into multimodal subtasks and refine the predicted bounding boxes progressively.

DetToolChain allows MLLMs to support various detection tasks without instruction tuning. Concretely, it significantly improves baseline GPT-4V and Gemini by \textbf{20\%-50\%} on main detection metrics of open-vocabulary detection, described object detection, referring expression comprehension, and oriented object detection. Compared to existing state-of-the-art methods, GPT-4V with our DetToolChain improves state-of-the-art object detectors by +\textbf{21.5\%} $AP_{50}$ on MS COCO \texttt{Novel class} set for open-vocabulary detection, \textbf{+24.23\%} $Acc$ on RefCOCO val set for zero-shot referring expression comprehension, \textbf{+14.5\%} $AP$ on D-cube describe object detection FULL setting. Notably, our method without trained on COCO \texttt{train2017} set even achieves better $AP_{50}$ on COCO \texttt{val2017} than the DETR trained on COCO \texttt{train2017} set. 
To summarize, our contributions are two-fold: (1) We propose a new prompting paradigm to instruct the MLLM to manage object detection by applying tools in the detection prompting toolkit with a multimodal detection Chain-of-Thought. (2) We propose detection prompting toolkits including \emph{visual processing} prompts and \emph{detection reasoning} prompts to facilitate MLLMs on detection tasks. We showcase that our method achieves remarkable performance on a range of detection tasks.

\section{Related Work}
\label{sec:related}

\noindent \textbf{Multimodal Large Language Models on Detection Tasks.}

The implementation of Multimodal Large Language Models (MLLMs) in detection tasks has been a recent focus. The widely used strategy is to finetune MLLMs using a high-quality image-text instruction tuning dataset consisting of detection problems. To promote the detection performance, BuboGPT~\cite{zhao2023bubogpt}, Kosmos-2~\cite{peng2023kosmos}, Shikra~\cite{chen2023shikra}, Qwen-VL~\cite{Bai2023QwenVLAF}, SPHINX~\cite{lin2023sphinx}, and Ferret~\cite{you2023ferret} constructed instruction datasets with high-quality question-and-answer pairs about detection. Consequently, they consumed great manual efforts and heavy computing costs and suffered from unsatisfied generalization ability to detect objects unseen in the instruction dataset. Furthermore, since most state-of-the-art MLLMs are closed-sourced~\cite{achiam2023gpt,team2023gemini,Bai2023QwenVLAF}, it is infeasible to apply instruction tuning on these MLLMs. In this paper, we believe the current state-of-the-art MLLM is potentially a zero-shot detector with proper promptings, and therefore design a Chain-of-Thought with new detection prompts to unleash their potential on detection tasks.

\noindent \textbf{Visual Prompting.}
Visual prompting, inspired by textural prompting in Natural Language Processing~\cite{raffel2020exploring, brown2020language}, manipulates images to improve the visual perception ability of MLLMs, such as classification and spatial reasoning. For example, RedCircle~\cite{shtedritski2023does} used a circle marker to guide the model to focus on the specific region for fine-grained classification while FGVP~\cite{yang2024fine}, SCAFFOLD~\cite{lei2024scaffolding} and SOM~\cite{yang2023setofmark} explored prompts for spatial reasoning with dot matrices or pretrained models such as SAM~\cite{kirillov2023segment}. Although CPT~\cite{yao2021cpt}, SOM~\cite{yang2023setofmark} are claimed to perform well on the visual grounding task and referring expression comprehension task, they just choose one of the pre-extracted bounding boxes and segmentation masks by high-quality detectors~\cite{yu2018mattnet} and segmentors~\cite{kirillov2023segment} for the correct answer, which is essential not the ability of object detection. To unleash the detection potential of MLLMs, we propose a general detection prompting toolkit that incorporates some well-adopted prior knowledge and strategies but does not include any pretrained model for detection or segmentation.

\noindent \textbf{Multimodal Chain-of-thought.}
Chain-of-Thought~\cite{wei2022chain, zhang2022automatic} methods and their variants~\cite{yao2024tree,besta2023graph,lei2023boosting,yao2023beyond} have greatly boosted the reasoning ability of large language models (LMMs). By encouraging LLMs to follow a human-like, step-by-step reasoning process, these methods decompose complex tasks into easier subtasks and solve them sequentially to get the final answer. 
However, the original CoT shows inferior performance~\cite{mitra2023compositional} on vision-language tasks, \emph{e.g.}, compositional reasoning. Therefore, recent researches pivoted towards developing multimodal Chain-of-Thought methods, \emph{e.g.,} Multimodal CoT~\cite{zhang2023multimodal}, DDCoT~\cite{zheng2024ddcot}, HoT~\cite{yao2023thinking} for multimodal reasoning, Chameleon~\cite{lu2024chameleon}, Compositional CoT~\cite{mitra2023compositional} for compositional reasoning, and Spatial CoT~\cite{chen2024spatialvlm} for spatial comprehension. Nevertheless, there remain multimodal Chain-of-Thought methods tailored for the detection ability of MLLMs unexplored. Our research fills this gap by introducing a detection-tailored Chain-of-Thought approach, coupled with a detection prompting toolkit. Our method differs from other practices by in visual promptings that use detection priors to progressively refine detection results through the CoT process.

\section{Method}
\label{sec:method}
While state-of-the-art MLLMs, \emph{e.g.}, GPT-4V and Gemini, exhibit commendable reasoning and recognition capabilities, their detection proficiency is still  \textbf{unleashed}. 
To unlock the potential of MLLMs in object detection, we introduce a comprehensive detection prompts toolkit (Sec.~\ref{sec:toolkit}) including \emph{visual processing} prompts and \emph{detection reasoning} prompts, and a detection Chain-of-Thought (Det-CoT) to reason the sequential implementation of the detection prompts in the toolkit (Sec.~\ref{sec:prompt_engineer}).

\subsection{Detection Chain-of-Thought for Object Detection}
\label{sec:prompt_engineer}
Let $\mathbb{L}$ and $\mathbb{I}$ denote the set of finite language strings and images, respectively. 
Given a test-time query $x =\{x_l, x_i | x_l \in \mathbb{L}, x_i \in \mathbb{I} \}$, which can be a specific detection task described in images and natural languages, we aim to obtain corresponding textual outputs with the help of a frozen $\mathtt{MLLM}$. 
This model, like GPT-4V~\cite{achiam2023gpt}, also maintains a prompt history that may include a list of previous messages, symbolized by $\mathcal{H}$, and produces a corresponding textual response $y$. The textual response $y$ includes the detection outputs, the diagnosis of current outputs and the suggestions of next prompts in the toolkit $\mathtt{TK}=\{T_{1}, \ldots, T_{n}\}$ to be called, where $T_i$ is the $i$-th tool pretrained in the toolkit. 
Finally, we define string extractors $e$, which is designed to retrieve a substring that is enclosed within specific delimiters.

\noindent \textbf{Algorithmic Procedure.} As illustrated in Algorithm~\ref{alg:main-alg}, we provide a conceptual overview of the procedure below: 
\begin{enumerate}

\item \textbf{Formatting the Test-time Query}: Using the transformation function $f$, the raw query is replaced with a suitable template for the MLLM.

\item \textbf{Loop Iteration in each prompting step}:
    \begin{enumerate}
    \item \textbf{Proposing a thought based on the response $y_{s}$}: The current message history, named $\mathcal{H}_s$, guides the MLLM to propose a thought - either to end it here and return the final outputs, or to continue the algorithm and engage another visual prompting tool.
    \item \textbf{Engaging detection prompting tool}: 
    If the MLLM does not return the result, it can select the appropriate visual prompting tool and construct it with corresponding instructions. The constructed prompting will be appended with the message history $\mathcal{H}_{s}$ and be sent to the next iteration.
    \end{enumerate}

\item \textbf{Returning the Final Response}:
If the string extractor $e$ discovers the final detection with special markers in the MLLM's response $y_s$, the algorithm will be ended and return the extracted bounding boxes. 
\end{enumerate}

\begin{algorithm}[t]
\caption{Detection Chain-of-Thought (Det-CoT)}
\textbf{Input}: $x=\{x_l,x_i\|x_l\in\mathbb{L}, x_i\in\mathbb{I}\}; thought\in\mathbb{L};~ \mathtt{TK}=\{T_{1}, \ldots, T_{n}\};~ \texttt{MLLM}:\mathbb{L},\mathbb{I}\rightarrow \mathbb{L}; f, e: \mathbb{L} \rightarrow \mathbb{L};~CoT~step~s;~string~concatenation~\oplus $
\begin{algorithmic}[1] %
\State $\mathcal{H}_{1} \gets f(x)$ \Comment{Format the test-time query to templated input}
\While{true} 
        \State $y_{s} \gets \mathtt{MLLM}~(\mathcal{H}_{s})$
        \State $thought_{s} \gets e(y_s)$ \Comment{Propose a thought in current step $s$}
        \If{$thought_{s}$ is not ending in this step} 
            \State $T_{n} \gets thought_{s} \times \mathtt{TK}$ 
            \Comment{Select the appropriate visual prompting tool}
            \State $\mathcal{H}_{s+1}\gets T_{n}(\mathcal{H}_{s})\oplus\mathcal{H}_{s} $  \Comment{Execute visual prompting tool}
        \Else  
            \State \Return{$e(y_{s})$} \Comment{Respond final output until all conditions are satisfied}
        \EndIf
        \EndWhile
\end{algorithmic}
\label{alg:main-alg}
\end{algorithm}

\subsection{Detection Prompting Toolkits}
\label{sec:toolkit}

\begin{figure}[t]
  \centering
  \includegraphics[width=\linewidth]{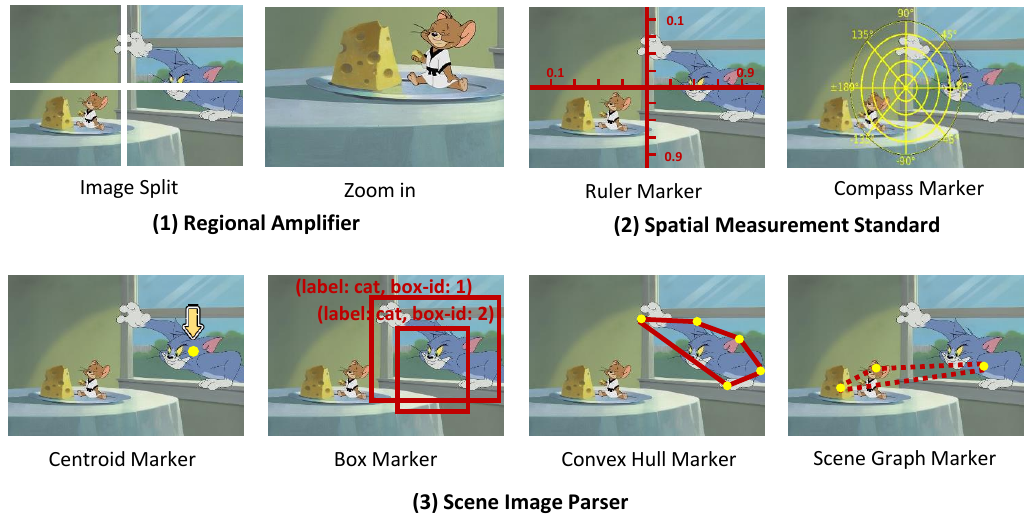}
  \caption{Illustration of visual processing prompts. We design Regional Amplifier, Spatial Measurement Standard, and Scene Image Parser to facilitate the detection ability of MLLMs from different perspectives.
  }
  \label{fig:vis_process}
\end{figure}

Compared with language promptings, visual promptings are more intuitive and can effectively teach MLLMs prior knowledge for detection. Following this insight, we design a comprehensive set of visual prompting toolkits to enhance various detection tasks. The toolkits are summarized in Fig.~\ref{fig:vis_process} and Tab.~\ref{tab:det reason} and described as follows. 

\begin{table}[t]
  \centering
  \small
  \tabcolsep=1em
  \caption{Detection Reasoning prompts and their effects.}
  \resizebox{\linewidth}{!}{
    \begin{tabular}{ll}
    \toprule
    Name  & Effect \\
    \midrule
    \multirow{1}[0]{*}{\texttt{Problem Insight Guider}} & Find potential detection problems in the image and provide advice. 
          \\
          \midrule
      \multirow{1}[0]{*}{\texttt{Spatial Relationship Explorer}} &  Leverage spatial reasoning ability to analyze the relationship of objects.
    \\
    \midrule
    \multirow{2}[0]{*}{\texttt{Contextual Object Predictor}} & Based on common sense, inquire the model what objects are likely \\
          &  to co-occur in the image. \\
\midrule
    \multirow{2}[0]{*}{\texttt{Self-Verification Promoter}} & Verify the detection results itself to ensure consistent predictions and \\
          & decide when to end the algorithm. \\
    \bottomrule
    \end{tabular}%
    }
  \label{tab:det reason}%
  \vspace{-2em}
\end{table}%

\subsubsection{Visual Processing Prompts.} We introduce \emph{visual processing} prompts to pre-process the input image to enhance the detection performance of MLLMs. Inspired by prior techniques for detection, we design our \emph{visual processing} prompts (see Fig.~\ref{fig:vis_process}) with three types focusing on different aspects, \emph{i.e.}, better visibility on details, more accurate spatial reference, and better contextual comprehension of the given images. 

\begin{enumerate}
\item\textbf{Regional Amplifier} aims at enhancing the visibility of the region of interest for MLLMs. Specifically, \texttt{Image Split} crops the image into disparate parts, and \texttt{Zoom in} enables close-up inspection of specific regions in the image.

\item\textbf{Spatial Measurement Standard} provides a more explicit reference for object detection by overlaying rulers and compasses with linear graduations on the original image, as depicted in Fig.~\ref{fig:vis_process}(2). The auxiliary ruler and compass enable MLLMs to output accurate coordinates and angles with the help of translational and rotational references overlaid in the image. Essentially, this standard simplifies the detection task, allowing MLLMs to read out the coordinates of the objects instead of directly predicting them. 

\item\textbf{Scene Image Parser} marks the predicted location or relations of objects, enabling a more localized understanding of images using spatial and contextual information. The proposed scene image parser can be divided into two categories. First, we mark the predicted objects with centroids, convex hulls and bounding boxes with label names and box indices. These markers represent the object location information in different formats, enabling the MLLM to detect diverse objects with different shapes and backgrounds, especially with irregular shapes or significant occlusions. For example, \texttt{Convex Hull Marker} marks the boundary points of objects and connects them as the convex hull to enhance the detection performance when objects have very irregular shapes. Second, we mark the scene graph by connecting the center of different objects with \texttt{Scene Graph Marker} to highlight the relationships of objects in the image. Based on the scene graph, the MLLM can leverage its contextual reasoning abilities to refine its predicted boxes and avoid hallucinations. For instance, as shown in Fig.~\ref{fig:vis_process}(3), Jerry Mouse is going to eat cheese, so their bounding boxes should be pretty close to each other.
\end{enumerate}

\begin{wrapfigure}{r}{0.335\textwidth}
  \centering
  \vspace{-2.4em}
  \includegraphics[width=0.335\textwidth]{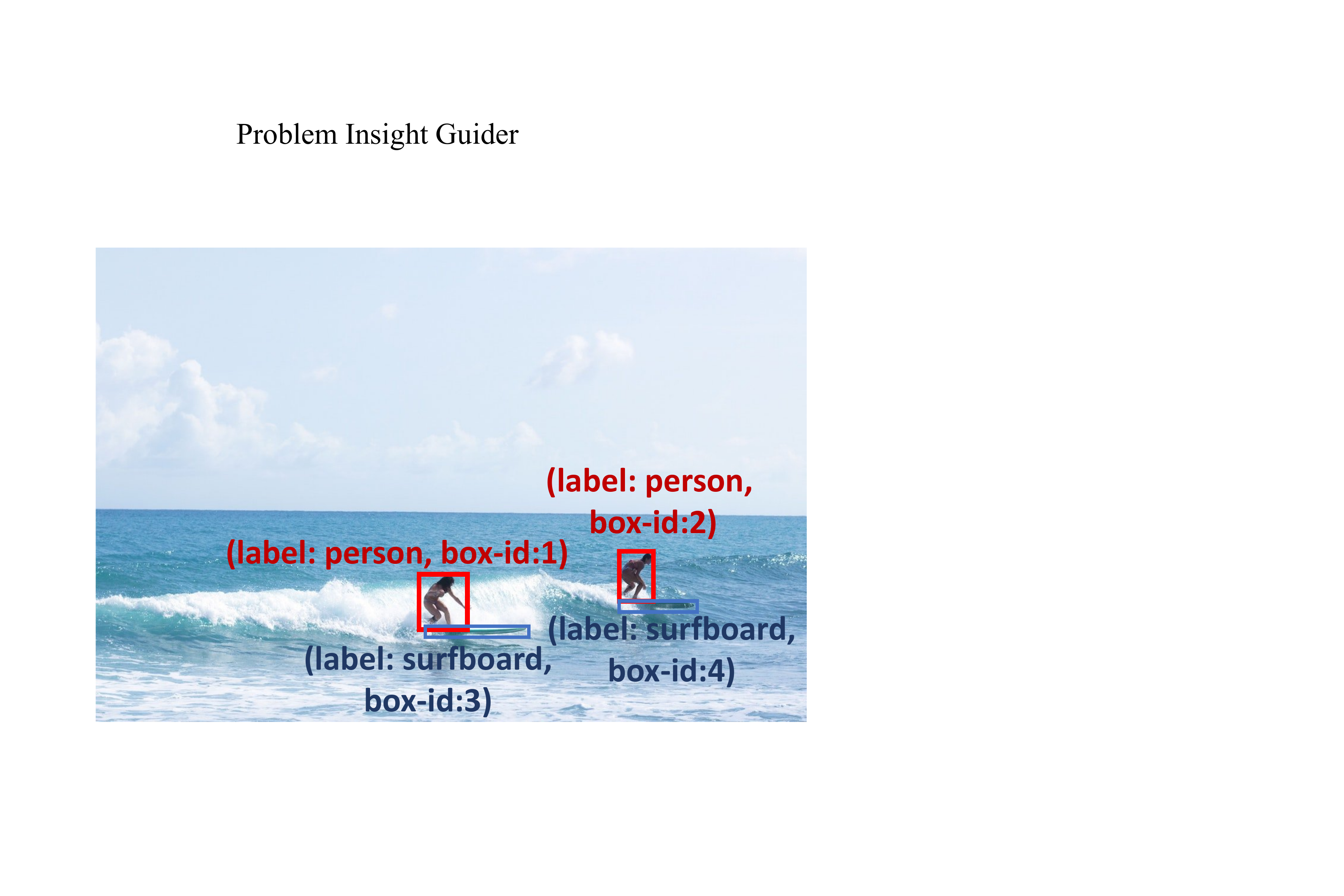}
  \caption{Detection reasoning prompts can help diagnose small object detection problems, e.g., using common sense to locate the surfboards under people's feet and encouraging the model to detect surfboards in the ocean.
  }
  \label{fig:det_reason}
  \vspace{-2.em}
\end{wrapfigure}

\subsubsection{Detection Reasoning Prompts.}
To improve the reliability of predicted boxes, we conduct \emph{detection reasoning} prompts (illustrated in Tab.~\ref{tab:det reason}) to examine the predictions and diagnose the potential problem that may exist. Based on these intuitions, we first propose \texttt{Problem Insight Guider}, which highlights difficult issues and offers effective detection suggestions for query images. 
For example, given Fig.~\ref{fig:det_reason}, the prompt will point out the issue of small object detection and suggest solving it by zooming in the region of the surfboard. Second, to leverage the inherent spatial and contextual ability of MLLMs, we design \texttt{Spatial Relation Analyzer} and \texttt{Contextual Object Preditor} to ensure the detection results accord with common sense. As shown in Fig.~\ref{fig:det_reason}, surfboards may co-occur with the ocean (contextual knowledge), and the surfing person should have a surfboard very close to his feet (spatial knowledge). Furthermore, we apply \texttt{Self-Verification Promoter} to enhance the consistency of multi-round responses.
To further promote MLLMs' reasoning ability, we engage in widely adopted prompting methods, \emph{e.g.}, debating~\cite{du2023improving,zhuge2023mindstorms} and 
self-debugging~\cite{schick2022peer,chen2023teaching}.

\begin{figure}[t]
  \centering
  \includegraphics[height=12.3cm]{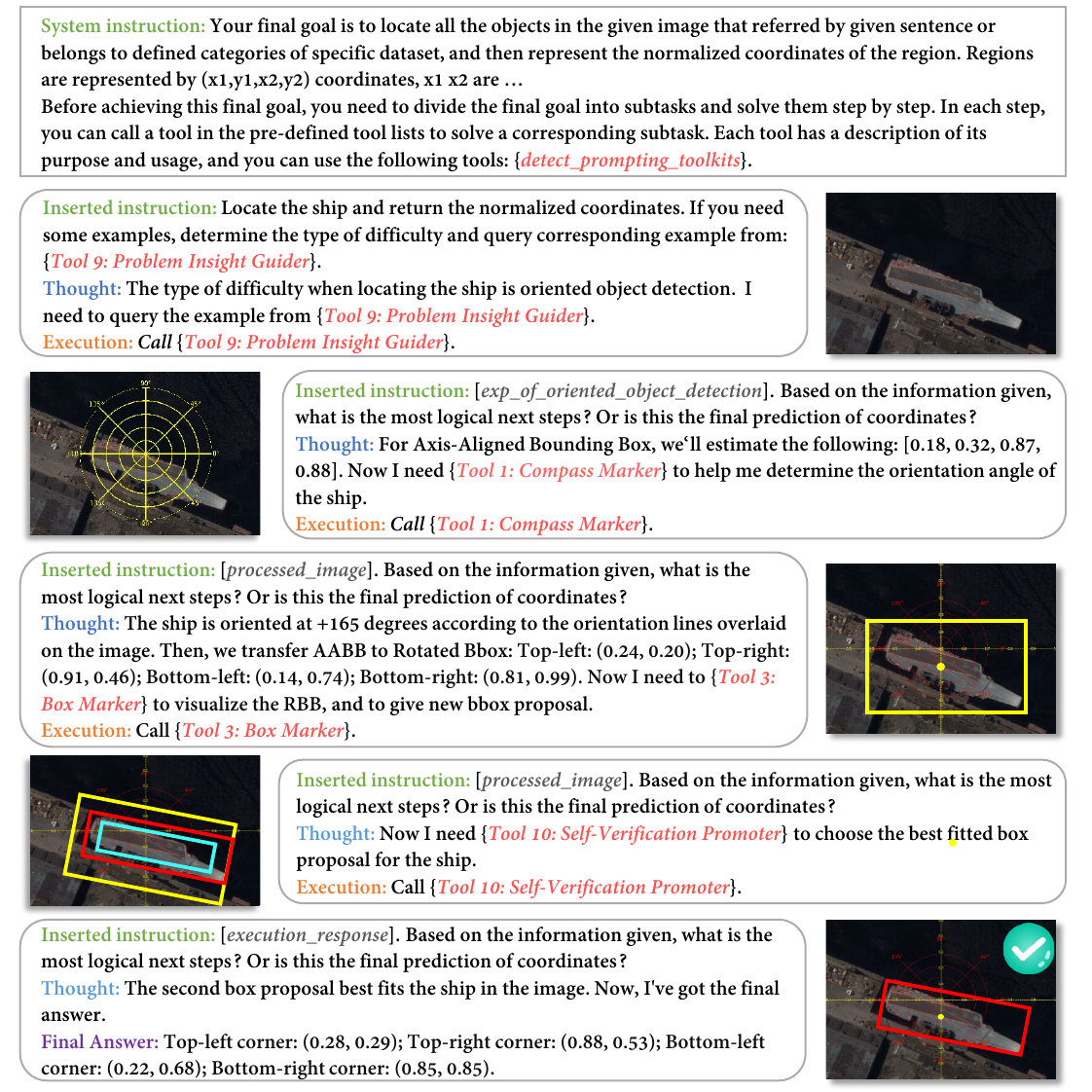}
  \caption{
  An example of DetToolChain for oriented object detection on HRSC2016.
  }
  \label{fig:vis-rotate}
\end{figure}

\section{Experiments}
\label{exp}
\subsection{Experimental Setup}
\noindent \textbf{Benchmarks.} 
To assess the effectiveness of our proposed methodology, we undertook a comprehensive evaluation across a diverse array of tasks and datasets related to object detection. We also extend our method to Referring expression comprehension which also needs to localize objects referred to in the expression. These evaluations encompass:
\begin{itemize}
\item \textbf{Object detection} utilize the MS COCO 2017 dataset~\cite{lin2014microsoft}, with results reported on the validation subset by default.
\item \textbf{Described object detection} on the D-cube dataset~\cite{xie2024described,wu2023advancing}, aiming to confirm the presence of objects described in arbitrary open-set expressions and to localize them accordingly.
\item \textbf{Referring expression comprehension}, a multimodal task that involves grounding a referent based on a given expression. This was tested on three representative benchmarks: RefCOCO~\cite{yu2016modeling}, RefCOCO+~\cite{yu2016modeling}, and RefCOCOg~\cite{mao2016generation}.
\item \textbf{Oriented object detection} on the widely recognized HRSC2016 dataset~\cite{liu2016ship}, aiming to identify and locate objects of arbitrary orientation within images.
\end{itemize}

\noindent \textbf{Models.}
We utilized the following Multimodal Large Language Models (MLLMs) in our experiments to explore the performance of our proposed DetToolChain:
\begin{itemize}
\item GPT-4~\cite{achiam2023gpt}, specifically the \texttt{gpt-4-vision-preview} model.
\item Gemini~\cite{team2023gemini}, specifically the \texttt{gemini-pro-vision} variant.
\end{itemize}

\noindent \textbf{Implementations.} 
As shown in Fig.~\ref{fig:vis-rotate}, the system instruction is set to ensure MLLM to comprehend its ultimate task objective and the preceding steps. Subsequently, the task description and input image are provided as the initial round of inserted instruction.  
Upon reaching the final answer, an automatic function extracts coordinate results, streamlining the prediction collection process.
To accommodate varying image sizes across datasets, we performed a transformation from absolute to normalized coordinate axes.
The returned normalized coordinates are ultimately converted back to absolute ones for standard evaluation.
For comparison, we implement baselines of GPT-4V and Gemini. We provide foundational prompt to enable successful result return. Across all experiments, we uniformly applied identical parameters and instructions, setting the maximum token count to 1024, with other settings at default values.

\subsection{Experimental Results}

\noindent \textbf{Open Vocabulary Object Detection}
As shown in Tab.~\ref{tab:ovd}, we evaluate our method on open vocabulary detection (OVD), where the $AP_{50}$ results on 17 novel classes, 48 base classes, and all classes of the COCO OVD benchmark~\cite{bansal2018zero} are reported. 
With our \name, the performance of GPT-4V and Gemini are both remarkably enhanced. GPT-4V+\name significantly outperforms the state-of-the-art method CORA~\cite{wu2023cora} by $\textbf{22.7} AP_{50}$, $\textbf{4.1} AP_{50}$, and $\textbf{9.0} AP_{50}$ on Novel, Base and All classes, respectively. 
Benefiting from general comprehension abilities, MLLMs can originally recognize most objects in the image, including those belonging to novel classes. However, they fall short in predicting precise textural coordinates. When integrated with our \name, the MLLM is instructed to locate the position with overlaying measure standards. 
This enhancement allows the MLLM to recognize and accurately locate objects across novel classes, effectively transforming it into a super open-vocabulary detector.

\begin{table}[t]
\parbox{.475\linewidth}{
\centering
\caption{Main results on the COCO OVD benchmark. We report $AP_{50}$.
}
\label{tab:ovd}
\resizebox{1.0\linewidth}{!}{
    \begin{tabular}{lccc}
    \toprule
    Methods & Novel & Base  & All \\
    \cmidrule(lr){1-1}\cmidrule(lr){2-4} 
    OCR-CNN~\cite{zareian2021open} &22.8 &46.0 &39.9 \\
    ViLD~\cite{gu2021open}  & 27.6  & 59.9  & 51.3  \\
    OV-DETR~\cite{zang2022open} & 29.4  & 61.0  & 52.7  \\
    MEDet~\cite{chen2022open} & 32.6  & 54.0  & 49.4  \\
    CFM-ViT~\cite{kim2023contrastive} & 34.1 & - & 46.0 \\
    OADP~\cite{wang2023object} & 35.6 & 55.8 & 50.5 \\
    RegionCLIP~\cite{zhong2022regionclip} & 39.3  & 61.6  & 55.7  \\
    BARON~\cite{wu2023aligning} & 42.7 & 54.9 & 51.7 \\
    CORA~\cite{wu2023cora}  & 43.1  & 60.9  & 56.2  \\
    \cmidrule(lr){1-1}\cmidrule(lr){2-4}
    Gemini (baseline)~\cite{team2023gemini} & 2.2   & 0.9   & 1.2  \\
    Gemini +DetToolChain & 63.6  & 58.5  & 59.8  \\
    GPT-4V (baseline)~\cite{achiam2023gpt} & 2.3   & 1.2   & 1.5  \\
    GPT-4V + DetToolChain & \textbf{65.8}  & \textbf{65.0}  & \textbf{65.2}  \\
    \bottomrule
    \end{tabular}%
}
}
\hfill
\parbox{.465\linewidth}{
\centering
\caption{Performance comparisons on COCO \texttt{val2017} for object detection.}
\label{tab:od}
\resizebox{1.0\linewidth}{!}{
    \begin{tabular}{lccc}
    \toprule
           Methods & \multicolumn{1}{c}{$AP$} & \multicolumn{1}{c}{$AP_{50}$} & \multicolumn{1}{c}{$AP_{75}$}\\
           \cmidrule(lr){1-1}\cmidrule(lr){2-4}  
           \textit{Supervised methods} & & & \\
          \cmidrule(lr){1-1}\cmidrule(lr){2-4}  
     Faster R-CNN-R50~\cite{ren2015faster}&36.4 &58.4 &39.1 \\
          Mask R-CNN-R50~\cite{he2017mask}  &37.3 &59.0 &40.2\\
           DETR-R50~\cite{carion2020end} & 42.0& 62.4 &44.2 \\
           VisionLLM-R50~\cite{wang2023visionllm}  & 44.6 &64.0 &48.1\\
           UNINEXT-R50~\cite{Yan2023UniversalIP} &51.3 &68.4 &56.2   \\
           Co-DETR-L~\cite{zong2023detrs} & 60.7 &78.5& 66.7 \\
          \cmidrule(lr){1-1}\cmidrule(lr){2-4}  
          \textit{Zero-Shot methods} & & & \\
          \cmidrule(lr){1-1}\cmidrule(lr){2-4}  
           Gemini (baseline)~\cite{team2023gemini} &0.2   & 1.1   & 0.0 \\
           Gemini+\name &  30.6  & 58.0  & 27.9       \\
            GPT-4V\footnotemark[1]
            (baseline)~\cite{achiam2023gpt} &0.3   & 1.4   & 0.0 \\
           GPT-4V+\name &  
            34.5  & 64.8  & 31.5    \\
          \bottomrule
    \end{tabular}}
    }
\end{table}

To position our method in previous state-of-the-art object detectors, we further compare our zero-shot detection performance with their detection performance by the model trained on COCO \texttt{train2017} in Tab.~\ref{tab:od}. While the baseline performance of GPT-4V and Gemini are only $AP$ is 0.2 and 0.3 respectively, there is a significant performance improvement after employing our \name, \emph{i.e.,} Gemini's $AP$, $AP_{50}$, and $AP_{75}$ increasing to 30.6, 58.0, and 27.9, and GPT-4V's $AP$, $AP_{50}$ and $AP_{75}$ increasing to 34.5, 64.8, and 31.5. 
Notably, GPT-4V with DetToolChain surpasses Faster R-CNN on $AP_{50}$ by {\textbf{6.4}}, and even achieves better performance than recent detectors, \emph{e.g.}, DETR-R50~\cite{carion2020end} and VisionLLM-R50~\cite{wang2023visionllm}, demonstrating its potential to be a general object detector in the real world. 
The higher $AP_{50}$ but lower $AP_{75}$ than state-of-the-art detectors, \emph{e.g.,} Mask R-CNN and DETR, indicates that our method enables MLLMs to detect most objects in a zero-shot manner but with less accurate box boundaries because we consider accurate boundaries involves annotation priors in each dataset and can only be regressed after learning the model on the training set.
\footnotetext[1]{We have carefully designed the prompts for detection. While AP is a very strict metric, GPT-4V baseline achieves 20.4 and 12.7 on two less strict metrics, miou-precision and miou-recall, respectively, showing GPT-4V baseline has some ability for detection.}

\begin{table} 
	\centering
\caption{Performance comparisons on the described object detection task. We report $AP$ as the evaluation metric.}
\label{tab:dod}
    \begin{tabular}{lccc}
    \toprule
    Methods & FULL  & PRES  & ABS \\
     \cmidrule(lr){1-1}\cmidrule(lr){2-4} 
    {OFA-L}~\cite{wang2022ofa} & 4.2   & 4.1   & 4.6  \\
    OWL-ViT-L~\cite{minderer2022simple} & 9.6   & 10.7  & 6.4  \\
    G-DINO-B~\cite{Liu2023GroundingDM} & 20.7  & 20.1  & 22.5  \\
    UNINEXT-H~\cite{Yan2023UniversalIP} & 20.0  & 20.6  & 18.1  \\
    {SPHINX-7B}~\cite{lin2023sphinx} & 10.6  & 11.4  & 7.9  \\
    OFA-DOD-B~\cite{xie2024described} & 21.6  & 23.7  & 15.4  \\
    FIBER-B~\cite{dou2022coarse} & 22.7  & 21.5  & 26.0  \\
     \cmidrule(lr){1-1}\cmidrule(lr){2-4} 
    Gemini (baseline)~\cite{team2023gemini} & 14.3  & 14.2  & 14.2  \\
    Gemini + DetToolChain & 35.9  & 35.9  & 35.8  \\
    GPT-4V (baseline)~\cite{achiam2023gpt} & 16.3  & 16.3  & 16.2  \\
    GPT-4V + DetToolChain & \textbf{37.2}  & \textbf{37.5 } & \textbf{36.2 } \\
    \bottomrule
    \end{tabular}%
    \vspace{-1em}
\end{table}

\noindent \textbf{Described Object Detection}
Described Object Detection~\cite{xie2024described} is a new task that can be regarded as a superset of open-vocabulary detection and referring expression comprehension. The dataset D-cube~\cite{xie2024described,wu2023advancing} in the task presents two primary challenges: 
(1) The presence of negative samples, where the object referred by sentences does not necessarily exist within the given image. This demands the model's capability to understand semantically incongruent concepts.
(2) The referring sentences are exceedingly complex, often containing the use of negation, such as "\textit{the person riding a horse who is \textbf{not} wearing a hat}", evaluating the model's comprehension of complex and lengthy sentences.
The experiment evaluates the model's ability across three categories: FULL, PRES, and ABS, which respectively mean the full descriptions (422 categories), presence descriptions (316 categories), and absence descriptions (106 categories).

As shown in Tab.~\ref{tab:dod}, with our \name method, both GPT-4V and Gemini significantly outperform existing approaches, \emph{e.g.,} GPT-4V + \name improve the best method, FIBER-B~\cite{dou2022coarse}, by 14.5, 16.0, and 10.2 $AP$ in the FULL, PRES, and ABS settings, respectively. 
Such improvements come from two aspects. First, Gemini and GPT-4V have stronger ability to comprehend detection-related descriptions, especially the negative concepts, and get rid of  unrelated concepts in both PRES and ABS settings than other models, \emph{e.g.}, OFA~\cite{wang2022ofa}, OWL-ViT~\cite{minderer2022simple}. Second, our proposed DetToolChain significantly unleash the ability of MLLMs to detect the described objects by prompts in the detection toolkit and the new detection Chain-of-Thought.

\begin{table}[t]
\footnotesize
\centering
\caption{Performance comparisons on the referring expression comprehension task. We report Top-1 Accuracy@0.5 (\%) as the evaluation metric.  }
\begin{tabular}{lcccccccc}
\toprule
\multicolumn{1}{c}{} & \multicolumn{3}{c}{RefCOCO} & \multicolumn{3}{c}{RefCOCO+} & \multicolumn{2}{c}{RefCOCOg} \\
\multicolumn{1}{l}{\multirow{-2}{*}{Methods}} & val & test-A & test-B & val & \cellcolor[HTML]{FFFFFF}test-A & test-B & val-u     
 & test-u \\ 
\cmidrule(lr){1-1}\cmidrule(lr){2-4}\cmidrule(lr){5-7}\cmidrule(lr){8-9}{\textit{Supervised sota}}\\ 
\cmidrule(lr){1-1}\cmidrule(lr){2-4}\cmidrule(lr){5-7}\cmidrule(lr){8-9}

\multicolumn{1}{l}{UNINEXT ~\cite{Yan2023UniversalIP}} &  {92.64} &  {94.33} &  {91.46}  &  {85.24} &  {89.63} &  {79.79} &  {88.73} &  {89.37} \\
\multicolumn{1}{l}{G-DINO-L ~\cite{Liu2023GroundingDM}} & 90.56 & 93.19 & 88.24 &  {82.75} &  {88.95} &  {75.92}  &  {86.13} &  {87.02} \\ 
\multicolumn{1}{l}{MiniGPT-v2-7B~\cite{chen2023minigpt}}  & 88.69 & 91.65 & 85.33 & 79.97 & 85.12 & 74.45& 84.44 & 84.66 \\
\multicolumn{1}{l}{Shikra-13B~\cite{chen2023shikra}} & 87.83 & 91.11 & 81.81& 82.89 & 87.79 & 74.41  & 82.64 & 83.16\\
\multicolumn{1}{l}{ SPHINX~\cite{lin2023sphinx}}  & 89.15 & 91.37 & 85.13& 82.77 & 87.29 & 76.85 & 84.87 & 83.65 \\
\multicolumn{1}{l}{Qwen-VL-7B~\cite{Bai2023QwenVLAF}}  & 89.36 & 92.26 & 85.34 & 83.12 & 88.25 & 77.21& 85.58 & 85.48 \\
\cmidrule(lr){1-1}\cmidrule(lr){2-4}\cmidrule(lr){5-7}\cmidrule(lr){8-9}{\textit{Zero-Shot methods}}\\ 
\cmidrule(lr){1-1}\cmidrule(lr){2-4}\cmidrule(lr){5-7}\cmidrule(lr){8-9}
\multicolumn{1}{l}{CPT-Seg~\cite{yao2021cpt}}   & 32.20  & 36.10  & 30.30    & 31.90  & 35.20  & 28.80& 36.70  & 36.50 \\
\multicolumn{1}{l}{CPT-adapted~\cite{subramanian2022reclip}} &23.79 &22.87 &26.03 &23.46 &21.73 &26.32  &21.77 &22.78\\
\multicolumn{1}{l}{GradCAM~\cite{Selvaraju2017gradcam}} &42.85 &51.07& 35.21 &47.83& 56.92 &37.70 &50.86 &49.70\\
    \multicolumn{1}{l}{ReCLIP~\cite{subramanian2022reclip}}  & 45.78 & 46.10  & 47.07  & 47.87 & 50.10  & 45.10 & 59.33 & 59.01 \\
        \multicolumn{1}{l}{StructureSimi~\cite{han2023zero}}  & 49.37 &47.76& 51.68 &48.89 &50.02& 46.86& 60.95 &59.99\\
\cmidrule(lr){1-1}\cmidrule(lr){2-4}\cmidrule(lr){5-7}\cmidrule(lr){8-9}
\multicolumn{1}{l}{GPT-4V (baseline)}  & 25.48 & 26.22 & 24.39& 10.59 & 18.23  & 8.87  & 14.26 & 15.42 \\
\multicolumn{1}{l}{GPT-4V + \name}   & \textbf{70.01} & \textbf{72.33} & \textbf{67.24} &\textbf{ 57.32} & \textbf{62.74 } &\textbf{ 52.28 } & \textbf{63.48 }& \textbf{64.54 } \\
\bottomrule
\end{tabular}
\vspace{-1em}
\label{table:ref}
\end{table}

\noindent \textbf{Referring Expression Comprehension}
To demonstrate the effectiveness of our method on referring expression comprehension, we compare our approach with other zero-shot methods on the RefCOCO, RefCOCO+, and RefCOCOg datasets (Tab.~\ref{table:ref}). First, our \name improves GPT-4V baseline by 44.53\%, 46.11\%, and 24.85\% on val, test-A, and test-B, respectively, which exhibits the best zero-shot referring expression comprehension performance on RefCOCO. 
Since the referring sentences in RefCOCO contain a substantial number of directional words, such as "\textit{right}" and "\textit{left}", 
our designed \emph{visual processing} prompts, \emph{e.g.},\texttt{Ruler Marker}, can help MLLMs to more effectively understand the spatial rationale and detect target objects. Furthermore, while other zero-shot methods~\cite{subramanian2022reclip,han2023zero} introduce an out-of-the-shelf object detector~\cite{yu2018mattnet} to generate proposals, our \name instructs MLLM to predict and refine the detection boxes with their inherent spatial and contextual ability, which is more simple and adaptable in real deployments. In addition, our method significantly narrows the gap between zero-shot methods and instruction-tuned MLLMs~\cite{chen2023shikra,lin2023sphinx,chen2023minigpt,Bai2023QwenVLAF}, indicating the promising direction of designing new multimodal promptings.

\begin{table}[t]
  \centering
  \caption{Ablation studies of our proposed visual prompting tools and comparison with other Chain-of-Thought methods. \texttt{Veri.}, \texttt{SG}, and \texttt{Co-Obj.} denote \texttt{Self-Verification Promoter}, \texttt{Scene Graph Marker}, and \texttt{Contextual Object Predictor}, respectively.  }
    \adjustbox{max width=\textwidth}{
    \begin{tabular}{lcccccccccccc}
    \toprule
    \multirow{3}[1]{*}{Methods} & \multicolumn{10}{c}{MS COCO val2017 subsets  } & \multicolumn{2}{c}{HRSC2016 test set } \\
    \cmidrule(lr){2-11} \cmidrule(lr){12-13} 
    & \multicolumn{2}{c}{Single object  } & \multicolumn{2}{c}{Multi-instances  } & \multicolumn{2}{c}{Multi-objects} & \multicolumn{2}{c}{Occlusion object} & \multicolumn{2}{c}{Small object} & \multicolumn{2}{c}{Oriented Object} \\
          & \multicolumn{1}{c}{mIoUp} & \multicolumn{1}{c}{mIoUr} & mIoUp & mIoUr & mIoUp & mIoUr & mIoUp & mIoUr & mIoUp & mIoUr & mIoUp & mIoUr \\
           \cmidrule(lr){1-1}\cmidrule(lr){2-3} \cmidrule(lr){4-5}\cmidrule(lr){6-7} \cmidrule(lr){8-9}\cmidrule(lr){10-11} \cmidrule(lr){12-13} 
    (a) Gemini (baseline) & 0.19  & 0.19  & 0.25  & 0.13  & 0.17  & 0.09  & 0.22  & 0.17  & 0.13  & 0.07  & 0.16  & 0.09  \\
    (b) GPT-4V (baseline) & 0.23  & 0.23  & 0.25  & 0.15  & 0.18  & 0.11  & 0.24  & 0.18  & 0.16  & 0.10  & 0.13  & 0.08  \\
               \cmidrule(lr){1-1}\cmidrule(lr){2-3} \cmidrule(lr){4-5}\cmidrule(lr){6-7} \cmidrule(lr){8-9}\cmidrule(lr){10-11} \cmidrule(lr){12-13} 
    (c) GPT-4V + \texttt{Ruler} & 0.52  & 0.52  & 0.40  & 0.29  & 0.30  & 0.21  & 0.43  & 0.25  & 0.22  & 0.18  & 0.32  & 0.23  \\
    (d) GPT-4V + \texttt{Compass} & 0.22  & 0.22  & 0.27  & 0.16  & 0.19  & 0.10  & 0.28  & 0.17  & 0.14  & 0.06  & 0.47  & 0.33  \\
    (e) GPT-4V + \texttt{Split} + \texttt{Zoom}  & 0.43  & 0.43  & 0.52  & 0.43  & 0.45  & 0.37  & 0.43  & 0.33  & 0.46  & 0.38  & 0.34  & 0.28  \\
    (f) GPT-4V + \texttt{Ruler} + \texttt{Split} + \texttt{Zoom} & 0.64  & 0.64  & 0.61  & 0.54  & 0.52  & 0.47  & 0.55  & 0.49  & 0.54  & 0.47  & 0.49  & 0.41  \\
    (g) GPT-4V + \texttt{Ruler} + \texttt{Convex} + \texttt{Veri.} & 0.69  & 0.69  & 0.59  & 0.54  & 0.49  & 0.43  & 0.70  & 0.61  & 0.48  & 0.42  & 0.51  & 0.44  \\
    (h) GPT-4V + \texttt{Ruler} + \texttt{SG} + \texttt{Relation} & 0.60  & 0.60  & 0.60  & 0.54  & 0.53  & 0.48  & 0.64  & 0.59  & 0.49  & 0.44  & 0.47  & 0.40  \\
    (i) GPT-4V + \texttt{Ruler} + \texttt{Co-Obj.} + \texttt{Veri.} & 0.66  & 0.66  & 0.54  & 0.50  & 0.51  & 0.48  & 0.59  & 0.53  & 0.52  & 0.48  & 0.41  & 0.33  \\

               \cmidrule(lr){1-1}\cmidrule(lr){2-3} \cmidrule(lr){4-5}\cmidrule(lr){6-7} \cmidrule(lr){8-9}\cmidrule(lr){10-11} \cmidrule(lr){12-13} 
    (j) GPT-4V + ZS-CoT & 0.25  & 0.25  & 0.27  & 0.18  & 0.19  & 0.11  & 0.33  & 0.23  & 0.17  & 0.12  & 0.16  & 0.09  \\
    (k) GPT-4V + FS-CoT & 0.37  & 0.37  & 0.28  & 0.20  & 0.22  & 0.10  & 0.38  & 0.22  & 0.18  & 0.13  & 0.34  & 0.28 \\
    (l) GPT-4V + MM-CoT & 0.29  & 0.29  & 0.38  & 0.29  & 0.32  & 0.24  & 0.45  & 0.33  & 0.22  & 0.19  & 0.29  & 0.24  \\
               \cmidrule(lr){1-1}\cmidrule(lr){2-3} \cmidrule(lr){4-5}\cmidrule(lr){6-7} \cmidrule(lr){8-9}\cmidrule(lr){10-11} \cmidrule(lr){12-13} 
    (m) Gemini + \name & 0.75  & 0.75  & 0.67  & 0.61  & 0.56  & 0.51  & 0.77  & 0.70  & 0.52  & 0.44  & 0.62  & 0.56  \\
    (n) GPT-4V + \name & \textbf{0.78 } &\textbf{ 0.78 } &\textbf{ 0.75}  & \textbf{0.69}  & \textbf{0.64}  & \textbf{0.60}  & \textbf{0.78}  &\textbf{ 0.72}  & \textbf{0.61}  & \textbf{0.57}  & \textbf{0.63 } & \textbf{0.59}  \\
    \bottomrule
    \end{tabular}}
  \label{tab:abl}%
  \vspace{-1em}
\end{table}%

\noindent \textbf{Oriented Object Detection}
\label{sec:oriented}
To investigate the efficacy of our \name in detecting rotated objects, we conducted experiments on the HRSC2016 test set. The images in the HRSC2016 dataset contain several ships of varying orientations. 
We employ $mIoU_{precision}$ and $mIoU_{recall}$ as performance metrics. 
$mIoU_p$ reflects the precision in bounding box coordinates prediction for objects that have been identified and $mIoU_r$ evaluate recall for the model's ability to recognize all targeted objects. 
As shown in the last two columns of Tab.~\ref{tab:abl}, our \name significantly improves the baseline results of Gemini and GPT-4 by 0.46 $mIoUp$ and 0.47 $mIoUr$ on Gemini, and 0.50 $mIoUp$ and 0.51 $mIoUr$ on GPT-4V. Furthermore, compared with other combinations of detection prompts in Tab.~\ref{tab:abl}, \texttt{Compass Marker} brings
the most substantial enhancements on oriented object detection, 0.34 $mIoUp$ and 0.25 $mIoUr$ performance gains over the GPT-4 baseline, which is consistent with our expectations. With a compass drawn by \texttt{Compass Marker} on the query image, it is easier for MLLMs to read the angle out, which will utimately improve the accuracy of ship angle predictions. 

\subsection{Ablation Study and Further Analysis}
\label{sec:abl}
To further analyze the individual improvement of our proposed visual prompting tools, we classify the samples in COCO \texttt{val2017} into several detection scenarios:
\begin{itemize}
\item Single: Images containing only a single object.
\item Multi-instances: Images containing multiple instances of a single category.
\item Multi-objects: Images containing multiple objects of various categories.
\item Occluded objects: Images with objects significantly occluded (occlusion area greater than 50\%).
\item Small objects: Images with objects smaller than 32x32 pixels.
\end{itemize}

\noindent \textbf{The comparison with other CoTs?}
In Tab.~\ref{tab:abl}, we compare our \name with other Chain-of-Thought approaches, including: (1) Zero-shot CoT (ZS-CoT), initially proposed by adding "\textit{Let's think step by step}" to the original prompt, is adapted for our detection task as "\textit{Let's detect and localize target objects step by step}". (2) Few-shot CoT (FS-CoT), which provides demonstrations with intermediate reasoning steps in the prompt. (3) Multimodal CoT (MM-CoT), which introduces a two-stage process: generating a rationale based on the input of image and text, and performing inference based on the generated rationale. 
Among them, MM-CoT works better in scenarios with multiple instances, multiple objects, and occluded objects. While MM-CoT has managed to grasp the spatial relationships within images to some extent, it has not been directly applied to the localization of each target. However, by considering the distinct characteristics of each image, our \name marks spatial relationships directly on the images and explicitly uses them to detect every target, thereby achieving superior performance.

\noindent \textbf{Which tool is the best to use?}
Tab.~\ref{tab:abl} shows the effectiveness of various detection prompting tools. 
It is observed in exp-c and exp-d that \texttt{Ruler Marker} and \texttt{Compass Marker} enable a more intuitive grasp of translational and rotational scale for MLLM, and therefore benefit general detection and oriented object detection, respectively. For small object detection, we have two effective tools: \texttt{Image Split + Zoom in} to help MLLM select the region of interest and focus on fine-grained information in this region  (exp-e); and \texttt{Contextual Object Predictor} to reduce the missing rate of detection by the commonsense in LMMs (e.g., books highly exist in the library) (exp-i).
For multi-instance and multi-object detections, \texttt{Scene Graph Marker} and \texttt{Spatial Relationship Explorer} can improve the detection performance by using the spatial rationale among multiple objects (exp-h). 
For occluded targets, leveraging \texttt{Convex Hull Marker} and \texttt{Self-Verification Promoter} can contour the irregular visible parts of objects, thereby referring accurate bounding boxes (exp-g). 
Moreover, our DetToolChain can achieve better performance by strategically applying a combination of different detection prompting tools (exp-m \& exp-n).

\section{Conclusion}
\label{conclusion}
We have presented DetToolChain, a groundbreaking prompting paradigm, to unleash the potential of multimodal large language models (MLLMs), like GPT-4V and Gemini, as zero-shot detectors.
By employing a multimodal detection Chain-of-Thought (Det-CoT) to manage the proposed \emph{visual processing} prompts and \emph{detection reasoning} prompts,
MLLMs are instructed to perceive objects by regional focus, prediction progressive refinement, and contextual inference.
Our approach not only improves interpretability and precision in detection tasks across a variety of challenging scenarios, including occluded, small, and oriented objects, but also sets new records in open-vocabulary detection, described object detection, and referring expression comprehension without instruction tuning. 

\bibliographystyle{splncs04}
\bibliography{egbib}

\section{Appendix}
\subsection{Detection Reasoning Prompts}
In Tab.~\ref{table:tools}, we list the detailed usage description of provided detection reasoning prompts. To improve the reliability of predicted boxes, we conduct detection reasoning prompts to examine the predictions and diagnose the potential problem that may exist. It serves as a great component alongside the visual processing prompts.

\begin{table}[t]
\centering
\caption{Usage description of Detection Reasoning prompts.}
\begin{tabularx}{\textwidth}{@{}Xp{0.8\textwidth}@{}}
\toprule 
\textbf{Tool Name} & \textbf{Usage} \\ 
\midrule 
\texttt{Problem Insight ~~Guider} & Find potential detection problems in the image and offer prompting examples that suggest solutions to specific difficulty.
It helps tailor the Chain-of-Thought process for particular challenges and input samples. Example: [text: "There exists the difficulty of small object detection when locating the object in given image. Now I will give you a demonstration on small object detection.", image: \{example\_image\}, text: "In this demonstration image, cat is a small object. We can locate the approximate area of the cat in the image and then zoom in this area to achieve accurate localization of the cat."].
\\
\midrule 

\texttt{Self- Verification Promoter} & Encourage the model to check the accuracy or consistency of its own responses, promoting reliability and trustworthiness in its output. Besides, decide when to end the algorithm.
Example: "Based on all history information given, what is the most logical next steps? Or is this the final prediction of coordinates?" 
\\
\midrule 

\texttt{Contextual Object Predictor} & Based on the commonsense, inquire what objects are likely to co-occur in the current image to address issues of object hallucination and missing objects. 
Example: "What is the scene in the image? Considering what we usually find in such a scene, can you identify which objects are most likely to be present based on commonsense? Please list the objects you would expect to see besides the \{object\_list\} you've detected in this image."
\\
\midrule 

\texttt{Spatial Relationship Explorer} & Leverage spatial reasoning ability to analyze the relationship of objects. 
Example: "Using your spatial reasoning abilities, can you describe the relationships between the \{object\_bbox\_list\} you've detected in this image? Think about how the bounding box (bbox) might be positioned in relation to one another, and any potential interactions or connections between them. Based on this spatial rationale, refine the predicted bbox of each object."
\\
\midrule 

\texttt{Debate Organizer} & Combine responses from multiple threads to decide on the best answer through a debating mechanism, enhancing decision-making quality by leveraging diverse viewpoints. 
Example: "You are part of an expert panel tasked with optimizing an object detection algorithm. The goal is to ensure that every target object in the image is accurately enclosed by a bounding box (bbox). Based on the historical information and different bbox predictions: \{prediction1, prediction2, ...\}, decide which prediction is the best. Make sure that every object has a corresponding bbox without omitting any objects or including nonexistent objects. A good bbox should snugly fit the object's contours, neither extending beyond its edges nor being too small to enclose it fully.”
\\
\midrule 

\texttt{Expertise Identifier} & Provide a system message to set the assistant's behavior, modifying its personality or giving specific instructions for task execution. Example: "You are a helpful assistant skilled in \{the\_target\_task\}.".
\\
\bottomrule 
\end{tabularx}
\label{table:tools}
\end{table}

\subsection{Evaluation Metrics}
We employ $mIoU_{precision}$ and $mIoU_{recall}$ as performance metrics for oriented object detection and ablation study in the main paper. 
$mIoU_p$ reflects the precision in bounding box coordinates prediction for objects that have been identified and $mIoU_r$ evaluate recall for the model's ability to recognize all targeted objects. The detailed calculated process is denoted as:
\begin{equation} 
    \small
    mIoU_p=\frac{\sum_{TP}\left(IoU\right)}{\sum_{TP}\left(IoU\right)+\sum_{FP}\left(IoU\right)},~~ mIoU_r=\frac{\sum_{TP}\left(IoU\right)}{\sum_{TP}\left(IoU\right)+\sum_{FN}\left(IoU\right)},
\end{equation}
where TP, FP, and FN denote true positives, false positives, and false negatives, respectively.

\subsection{Detailed Prompts for Baseline method}
In our experiments, we observed that directly prompting GPT-4V to return precise coordinates of target objects resulted in a high likelihood of refusal to respond. To ensure fairness in experimental outcomes, we provided both GPT-4V and Gemini with basic prompts that enabled them to return in validated format. The prompt is: \textit{"Locate the referred object and represent the location of the region. Regions are represented by (x1,y1,x2,y2) coordinates. x1 x2 are the left and right most positions, normalized into 0 to 1, where 0 is the left and 1 is the right. y1 y2 are the top and bottom most positions, normalized into 0 to 1, where 0 is the top and 1 is the bottom."}.
\begin{figure}[t]
  \centering
  \includegraphics[height=11cm]{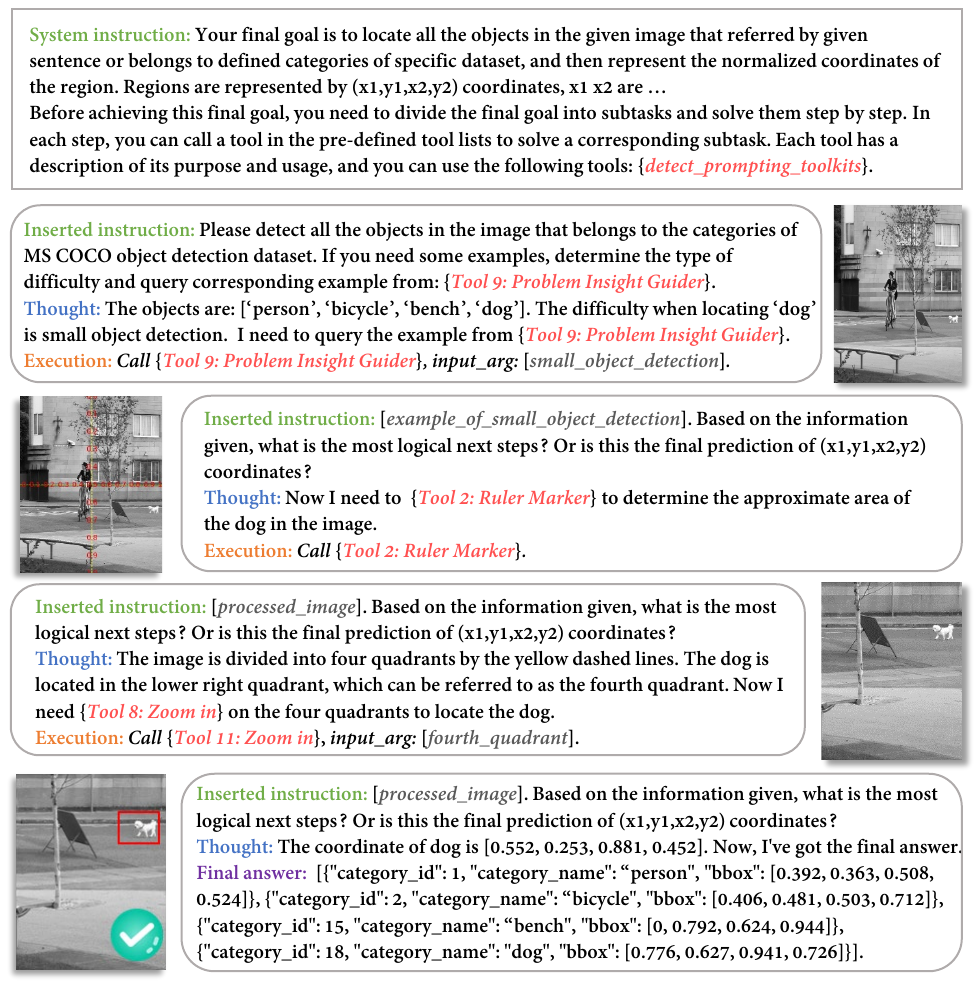}
  \caption{An example of DetToolChain for the recognizing process of multi-objects detection on MS COCO2017.
  }
  \label{fig:example1}
\end{figure}

\begin{figure}[t]
  \centering
  \includegraphics[height=8cm]{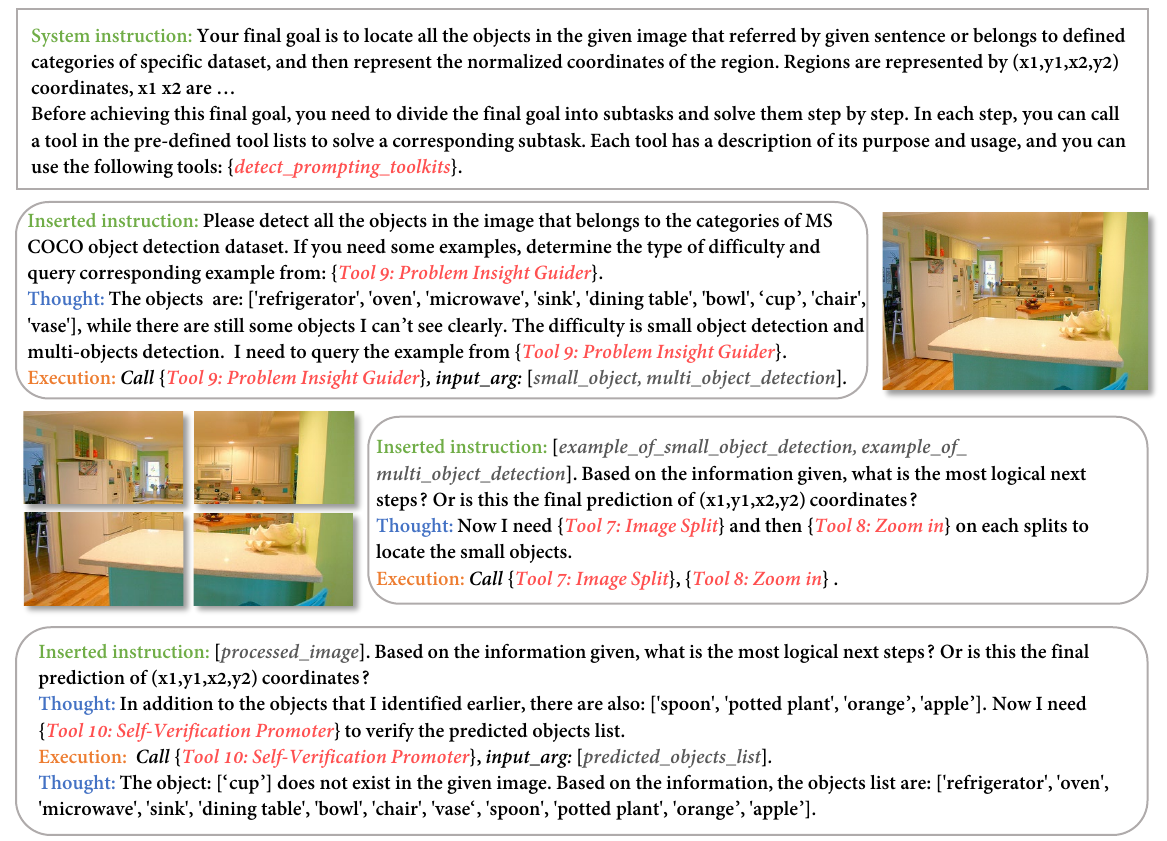}
  \caption{An example of DetToolChain for small object detection on MS COCO2017.
  }
  \label{fig:example2}
\end{figure}

\subsection{More examples of Det-CoT Process}
As shown in Fig.~\ref{fig:example1} and Fig.~\ref{fig:example2}, we provide more examples of detailed Det-CoT process.

\subsection{Limitations}
Despite the DetToolChain framework's pioneering approach to leveraging the zero-shot object detection capabilities of multimodal large language models (MLLMs), it encounters certain constraints that merit consideration. These limitations, while not undermining the overall efficacy and innovative nature of our method, highlight areas for future enhancement and optimization. 

\noindent \textbf{Operational Efficiency.} A primary constraint of the DetToolChain framework lies in its operational design, which adopts a sequential strategy. The inherent sequential processing of detection prompts, contingent on the outcomes of preceding operations, precludes the possibility of parallel computation. This linear dependency potentially affects the framework's overall speed and operational efficiency, signaling an area ripe for future innovation to enhance processing agility.

\noindent \textbf{Response Patterns.} 
The DetToolChain requires inputs in a specific format to work properly with the model API. However, it's not guaranteed that the outputs will always be in the expected format, like being enclosed in triple quotes in our framework.
This variability can lead to inconsistencies in multi-turn dialogues, where precise output formats are crucial for seamless interaction. Furthermore, the MLLM frequently incorporates apologetic statements within its response patterns, particularly in instances of suboptimal task performance, i.e., "\textit{Apologies for the confusion in my previous response}". This pattern of behavior is likely a result of the model's training on instruction-following data.

\noindent \textbf{Scalability.} The framework's reliance on extensive message histories and the need for a large-scale, contextually adept MLLM poses another limitation. While GPT-4V fulfills this requirement with its capacity for handling and retaining substantial textual information, smaller models like ChatGPT may not exhibit the same proficiency. This discrepancy underscores the necessity for a MLLM of significant scale and contextual understanding, a criterion not universally met across all existing language models.

\noindent \textbf{Cost Implications.} Engaging multiple rounds of interaction with the MLLM incurs elevated costs, a consideration of paramount importance given the resource-intensive nature of operating advanced MLLMs. Nonetheless, it is anticipated that these costs will diminish over time with the decreasing expenses associated with MLLM utilization.

In summary, while the DetToolChain framework presents a series of challenges related to operational efficiency, model API response patterns, the scalability and cost implications of the MLLMs, these limitations do not detract from the framework's overall value and effectiveness. Instead, they offer avenues for further research and development, aiming to refine and expand the utility of this innovative approach to object detection within the realm of MLLMs.

\subsection{Potential Negative Impact}
\noindent \textbf{Misuse in Surveillance.} The application of DetToolChain framework in surveillance technologies could potentially enhance the efficiency of mass surveillance systems, raising ethical concerns regarding civil liberties and individual privacy.

\noindent \textbf{Automated Decision Making.} The reliance on DetToolChain for critical decision-making in areas like autonomous driving, healthcare, or security could have severe implications if the MLLM's predictions are erroneous, leading to potential harm or unfair treatment.

\noindent \textbf{Economic Disruption.} The introduction of highly efficient detection systems might disrupt job markets, especially in sectors reliant on human visual inspection tasks, potentially leading to job displacement.

\noindent \textbf{Environmental Impact.} The significant computational resources required for running advanced MLLMs in DetToolChain framework contribute to carbon emissions and energy consumption, exacerbating the environmental footprint of large-scale AI research and deployment.

\subsection{Some Failure Cases}
\begin{figure}[h]
  \centering
  \includegraphics[height=8cm]{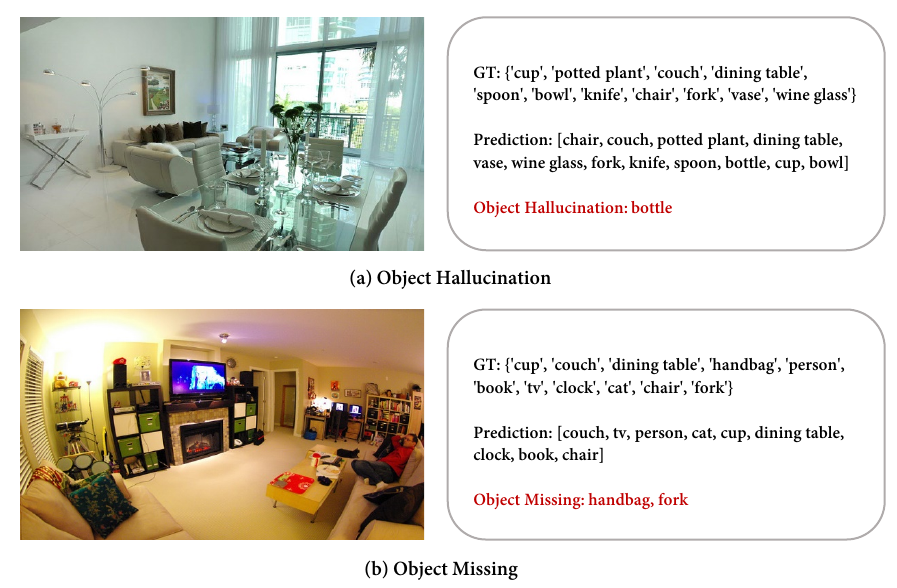}
  \caption{Failture cases of DetToolChain for the recognizing process of object detection on MS COCO2017.
  }
  \label{fig:example3}
\end{figure}

\begin{figure}[h]
  \centering
  \includegraphics[height=8.54cm]{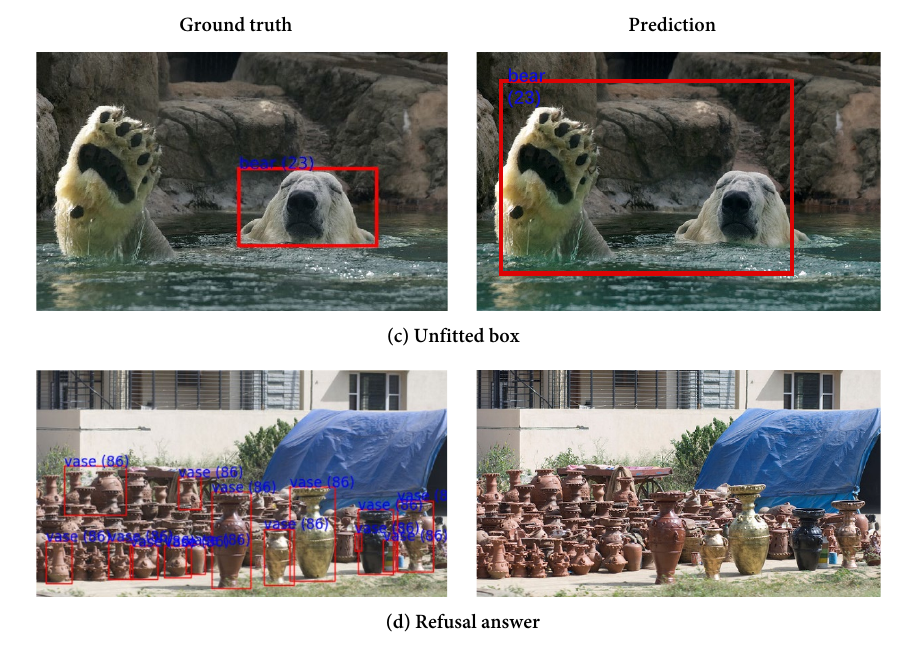}
  \caption{Failture cases of DetToolChain for the localizing process of object detection on MS COCO2017.
  }
  \label{fig:example4}
\end{figure}

Fig.~\ref{fig:example3} and Fig.~\ref{fig:example4} illustrate representative failure cases in our DetToolChain framework when processing the MS COCO2017 dataset. Common failures can be categorized into four types: (a) Object Hallucination, (b) Object Missing, (c) Unfitted Box, and (d) Refusal to Answer. 

\noindent \textbf{(a) Object Hallucination} refers to the incorrect prediction of object categories that do not actually exist in the image. These hallucinated objects often appear in conjunction with actual objects in the image, leading the model to erroneously infer their presence based on commonsense. 

\noindent \textbf{(b) Object Missing} denotes the omission of actual objects present in the image, which are often small and not prominently displayed. 

\noindent \textbf{(c) An Unfitted Box} is characterized by bounding box predictions that do not accurately encapsulate the object itself. This is due to our method not being fine-tuned on the MS COCO detection dataset, resulting in a lack of precise edge detection for objects. 

\noindent \textbf{(d) Refusal Answer} describes the model's refusal to respond to particularly challenging samples, such as the example given in Fig.~\ref{fig:example4}(d), where our method can correctly identify the sole presence of the category 'vase' yet refuses to provide the counts of vases or the bounding box for each. However, it is noteworthy that the ground truth provided by the MS COCO dataset is also not entirely accurate and fails to correctly identify all the bounding boxes for the vases.

\end{document}